\documentclass[conference, a4paper]{IEEEtran}
\IEEEoverridecommandlockouts
\usepackage{cite}
\usepackage{amsmath,amssymb,amsfonts}
\usepackage{algorithmic}
\usepackage{graphicx}
\usepackage{textcomp}
\usepackage{xcolor}
\def\BibTeX{{\rm B\kern-.05em{\sc i\kern-.025em b}\kern-.08em
    T\kern-.1667em\lower.7ex\hbox{E}\kern-.125emX}}

\renewcommand{\baselinestretch}{1.04} 

\begin{document}

\title{
Estimation of Individual Device Contributions for Incentivizing Federated Learning\\
\thanks{This work is supported in part by the US NSF under Grant ACI-1541069, JST PRESTO Grant no. JPMJPR1854, and KDDI fundation.}
}

\author{
\IEEEauthorblockN{Takayuki Nishio}
\IEEEauthorblockA{\textit{Graduate School of Informatics,} \\
\textit{Kyoto University,}\\
Kyoto, Japan \\
nishio@i.kyoto-u.ac.jp}
\and
\IEEEauthorblockN{Ryoichi Shinkuma}
\IEEEauthorblockA{\textit{Graduate School of Informatics,} \\
\textit{Kyoto University,}\\
Kyoto, Japan \\
shinkuma@i.kyoto-u.ac.jp}
\and
\IEEEauthorblockN{Narayan B. Mandayam}
\IEEEauthorblockA{\textit{WINLAB,} \\
\textit{Rutgers University,}\\
NJ, USA \\
narayan@winlab.rutgers.edu}
}

\maketitle

\begin{abstract}
    Federated learning (FL) is an emerging technique used to train a machine-learning model collaboratively using the data and computation resource of the mobile devices without exposing privacy-sensitive user data. 
    Appropriate incentive mechanisms that motivate the data and mobile-device owner to participate in FL is key to building a sustainable platform for FL. However, it is difficult to evaluate the contribution level of the devices/owners to determine appropriate rewards without large computation and communication overhead.
    This paper proposes a computation-and communication-efficient method of estimating a participating device's contribution level. The proposed method enables such estimation during a single FL training process, there by reducing the need for traffic and computation overhead. The performance evaluations using the MNIST dataset show that the proposed method estimates individual participants' contributions accurately with 46--49\% less computation overhead and no communication overhead than a naive estimation method.
\end{abstract}


\section{Introduction}
The expanded use of machine learning (ML) has empowered a wide variety of Internet-of-Things (IoT) applications, including fine-grained road-traffic and pedestrian prediction, fine-grained environment prediction, anomaly detection in network systems, and fraud detection in financial transactions \cite{mahdavinejad2018machine, cui2018survey}. Generally, ML requires tremendous computational power to quickly produce the analytical results. The steady growth of cloud computing platforms (e.g., AWS, Microsoft Azure, and Alibaba) has supported `resource-heavy' ML computing.

Data used by ML for IoT applications are generated by the IoT devices themselves, which are located at the network edge and include sensor devices, smartphones, and smart vehicles. People engaged in this activity have found this scenario to be problematic for the following two reasons. First, individuals and private organizations are concerned about the privacy of their data when sharing them with an entity that operates an IoT application. After the EU adopted its general data protection regulation (GDPR) \cite{voigt2017eu}, privacy became a top-prioritized issue with regard to IoT applications. The second reason concerns bandwidth costs, which individuals or private organizations must bear to support their IoT needs. As the volume of data increases, such bandwidth costs will become more problematic \cite{shinkuma2019data, inagaki2019prioritization}.
 
Federated learning (FL) was invented to tackle both of the above two issues \cite{McMahan16}. With traditional ML, the training dataset is usually stored at a central entity. Data must be first collected from the data sources to facilitate the learning process. FL focuses on data integration methods that comply with privacy and security laws. Under FL, data owners employ privacy protection techniques (e.g., homomorphic encryption, secret sharing, and differential privacy) to contribute model parameters trained on their own datasets to a federation. The federation then combines these local model parameters to train a more effective collective ML model. This allows the learning process to leverage the computational power of the data sources to train the model, similar to crowdsourcing.

Although FL has shown great advantages in enabling collaborative learning while protecting data privacy, it still faces an open challenge of incentivizing owners of IoT devices to join the FL effort by contributing their computation power and data \cite{zhan2020learning}. An intuitive idea is to reward participants according to their contributions. However, it is difficult to accurately evaluate their contributions. It has been reported that the relationship between model accuracy and the amount of training data is nonlinear. The model accuracy depends on model complexity and data quality. This accuracy can hardly be predicted in advance. 
%
Generally, in FL, data are unbalanced and non-i.i.d. between clients: as the training data present on the individual clients is collected by the clients themselves based on their local environment and usage pattern, both the size and the distribution of the local data sets will typically vary heavily between different clients \cite{sattler2019robust}.
%
Therefore, accurate estimation of individual contribution levels with small computation and communication overhead is a key to the success of incentivizing participants in FL.

This paper proposes a method that estimates the individual contribution level of FL participants with no overhead traffic and little computation overhead. Conventionally, a direct and accurate way of estimating their individual contribution levels is to first measure the degradation of model accuracy by removing the local model provided by each participant. This method is more computation-and communication-resource consuming, because it must repeat entire FL processes as many times as the number of participants. The proposed method achieves estimation during a single FL process with small computation overhead and no overhead traffic.
This is helpful for the incentive mechanism of FL when quickly determining rewards for each participant on the basis of their contribution levels. Of course, accuracy of the estimation must be also ensured. In this paper, performance evaluation using a real dataset is conducted to validate the proposed method in terms of estimation accuracy.

\section{Prior works}
%
Several studies on FL incentive mechanisms were published in 2019 and 2020, suggesting that this research topic is building momentum. Several basic analyses have been presented \cite{pandey2020crowdsourcing, lee2020market, kang2019design, chen2020mechanism}. 
Pandey et al. designed a framework in which FL-participating clients iteratively solved the local learning subproblems to meet an accuracy level that was subject to an offered incentive \cite{pandey2020crowdsourcing}. A communication-efficient cost model for the participating clients was established to formulate the incentive mechanism and to induce the necessary interaction between the mobile-edge computing (MEC) server and the participating FL clients. Introducing a two-stage Stackelberg game, they analyzed the game's equilibria and the response behaviors of the participating clients.
Lee et al. designed a distributed learning resource management mechanism over multiple MECs owned by different operators \cite{lee2020market}. They also presented a game theoretic approach that focused on analyzing the market behaviors and the economic benefits of FL by formulating and analytically solving a Stackelberg game model.
Kang et al. adopted contract theory to design an efficient incentive mechanism that mapped contributed resources into appropriate rewards to entice mobile-device owners possessing high-quality data to join FL and overcome information asymmetry issues \cite{kang2019design}. They presented the problem formulation and the optimal contract design of their mechanism and showed its superiority using the Stackelberg game model.
Chen et al. modeled and formulated the mechanism-design problem with type-imposed externalities \cite{chen2020mechanism}. For quasi-monotone externality-setting, they provided a revenue-optimal and truthful mechanism. For the general valuation functions, they provided both necessary and sufficient conditions for all truthful and individually rational mechanisms.

The method of determining rewards for clients is a key issue for FL incentive mechanisms \cite{kang2019mechanism, liu2020fedcoin, toyoda2019mechanism}. Kang et al. applied reputation as the necessary metric needed to assess the reliability of an FL-worker candidate, thus ensuring reliable worker selection. High-reputation workers bring high-quality data (i.e., high accuracy and reliability) to model training, generating reliable local model updates for FL tasks.
Liu et al. suggested that, to properly incentivize data owners to contribute their efforts, the Shapley value (SV) was often adopted to fairly assess their contribution \cite{liu2020fedcoin}. To help FL systems compute SVs to support sustainable incentive schemes, they proposed a blockchain-based peer-to-peer payment system: FedCoin. The SV of each FL client reflected its contribution to the global FL model in a fair way. It was calculated using the Proof-of-Shapley  consensus protocol, which replaced the traditional hash-based protocol in existing blockchain proof-of-works method.
Toyoda and Zhang introduced a competitive model-update method so that any rational worker who followed the protocol could maximize their profits. Each worker chosen in a certain round selects the top model updates submitted by workers in the previous round and updates their own model. Here, workers' reward is decided by the vote of the next round of workers. The motivation of choosing the models that achieve the best k models is that their model updates will have a greater chance to be voted for in the next round, meaning that more rewards will be obtained. This design will naturally compel rational workers to behave honestly without any heavy cryptography or special hardware.

Sustainability and fairness are also key issues of FL incentive mechanisms \cite{yu2020fairness, yu2020sustainable, zhan2020learning}.
Yu et al. considered a method to quantify the payoff for each data owner in order to achieve long-term systemic wellbeing \cite{yu2020fairness, yu2020sustainable}. Participants must incur some cost for contributing to the FL model with their local datasets. The training and commercialization of the models take time. Thus, there are delays before the federation has a sufficient budget to pay the participants. They further addressed this temporary mismatch between contributions and rewards.  
To address the issues of unshared decisions and ambiguous contribution evaluations, Zhan et al. designed an algorithm based on deep-reinforcement learning (DRL), which can learn system states from historical training records \cite{zhan2020learning}. DRLs can adjust the strategies of the parameter server and edge nodes according to environmental changes that may impose different requirements on training data.

Incentive mechanisms for FL in heterogeneous resource environments have also been studied \cite{lim2020hierarchical, jiao2020toward, zeng2020fmore}.


To the best of our knowledge, no prior work has addressed the reduction of computation and communication overhead for estimating the individual contribution levels of FL participants.

\section{Proposed method}
\subsection{System model}
\begin{figure}[t]
\centering
\includegraphics[width=1.0\linewidth]{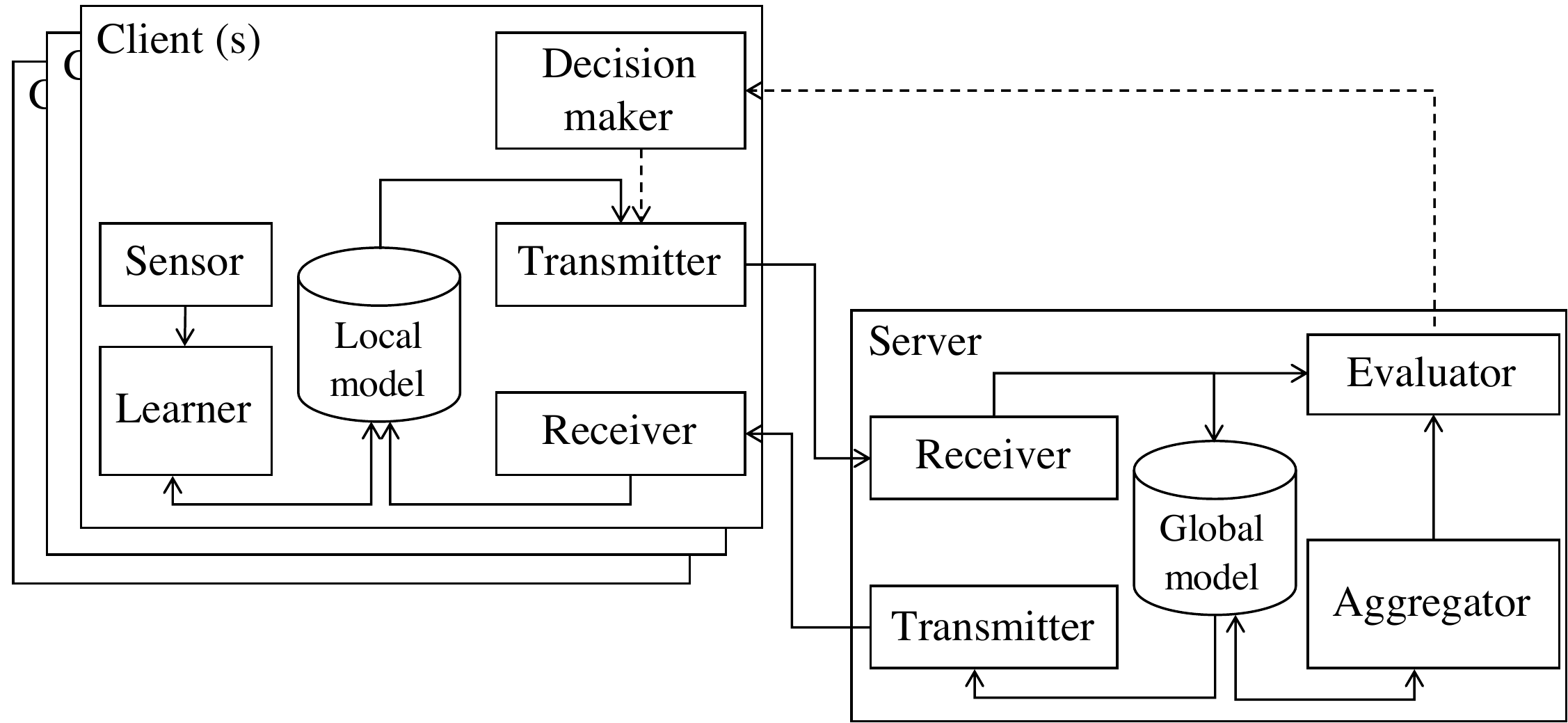}
\caption{System model}
\label{fig:system}
\end{figure}

Figure \ref{fig:system} shows our system model. The system comprises a server and multiple FL clients. At each client, the sensor generates data to be used for ML. The receiver receives an ML model from the server and stores it. The learner performs the training process to update the ML model using the data obtained from the sensor. The transmitter uploads the updated ML model to the server. The server ultimately receives the updated ML models from all the clients and stores them. The aggregator performs the model aggregation to update the ML model using the ML models obtained from the clients. The evaluator estimates the individual contribution level of each client by comparing the ML models before and after being updated using the clients' models. The evaluator sends back the rewards to each client based on its contribution. 
Note that decision making by clients is out of the scope of this paper and will be included in future work.
The decision maker at each client makes a decision to participate in FL if the reward is sufficient to compensate the cost of contribution of a client to the FL. Otherwise, a client leaves the FL.

\subsection{Design metrics to evaluate client contributions}\label{sec:metric}
Evaluating the contributions of each client accurately is not easy, because it is difficult to explicitly describe the actual improvement of model performance based on the quality of the clients' data and model (e.g., data amount, data variety, noisy label, the number of epochs, and noisy gradients, etc). Therefore, we next discuss model-performance based metrics that can be used to estimate the contributions of each client based on trained model performance.

Let $\mathcal{C} = {1,...,c}$ be a set of indices that describes $c$ clients. The amount of data samples possessed by client $i$ and the data distribution are denoted as $d_i$ and $v_i$, respectively. 
$M^{(r)}_i$ denotes the model updated by client $i$ at round $r$, and $M^{(r)}_\mathcal{S}$ denotes the model aggregated by those updated by a subset of clients, $\mathcal{S} \subseteq \mathcal{C}$. $M^{0}$ is defined as an initial model the parameters that are randomly determined. $P(\mathcal{M})$ is the performance score of model $M$, and the validation accuracy or loss can be used for the score.

In each round, each client updates $M^{(r-1)}_\mathcal{C}$ with its own data and uploads an updated model $M^{(r)}_i$. The server then aggregates the models into $M^{(r)}_\mathcal{C}$. The round is repeated until the $P(M^{(r)}_\mathcal{C})$ or $r$ achieves a certain threshold.

\vspace{3mm}
\noindent \textbf{Naive contribution metric } \\
Let $M_\mathcal{S}$ be a global model at the end of the FL where only clients in $\mathcal{S}$ participate in.
We define a naive metric for evaluating contributions of client $i$ as the gain in performance when the client joins the FL. 
The gain can modeled as the relative performance of a model that includes all clients to that which excludes only client $i$. Specifically, the normalized relative gain is given as:
\begin{equation}
    G^{\mathrm{Naive}}_i = \frac{P(M_\mathcal{C}) - P(M_{\mathcal{C} \setminus \{ i \}})}{\sum_{i \in \mathcal{C}}P(M_\mathcal{C}) - P(M_{\mathcal{C} \setminus \{ i \}})},
\end{equation}
where $P(M_{\mathcal{C} \setminus \{i\}})$ indicates a performance score in which all the clients excluding client $i$ train the model. The naive metric is intuitive, reasonable, and calculable.
However, calculating the metric requires additional FL training to obtain $M_{\mathcal{C} \setminus \{ i \}}$, which causes a large computational overhead and traffic overhead. Specifically, the computation and traffic overhead of calculating the naive metric can be given as:
\begin{eqnarray}
    \mathrm{Comp.:}&\; & \sum_{i\in\mathcal{C}} r_\mathrm{end} (C_\mathrm{server} + \sum_{j \in \mathcal{C}\setminus \{i\}}C^{(j)}_\mathrm{ud}), \label{eq:comp_naive}\\
    \mathrm{Traffic:}&\; & \sum_{i\in\mathcal{C}} r_\mathrm{end} \sum_{j \in \mathcal{C}\setminus \{i\}}2\Theta^{(j)}_\mathrm{model},\label{eq:traffic_naive}
\end{eqnarray}
where $r_\mathrm{end}$ is the number of rounds at the end of the FL task. $C^{(j)}_\mathrm{ud}$ and $C_\mathrm{server}$ denote the amount of computations for model update, namely stochastic gradient decent (SGD) operations, at client $j$ and the computation at the server, which is model aggregation and model validation. $\Theta^{(j)}_\mathrm{model}$ denotes traffic for transmitting a model from/to client $j$. The usual FL process causes computations of model aggregation and model updates at each client, the sum of which is $C_\mathrm{server} + \sum_{j \in \mathcal{C}}C^{(j)}_\mathrm{ud}$, at each round. Traffic is also generated for distributing and gathering models to/from FL clients at each round. The round is repeated $r_\mathrm{end}$ times. For calculating the naive metric for client $i$, the FL process with $\mathcal{C}\setminus \{i\}$ must be conducted individually, and thus the additional computation and traffic are as shown in \eqref{eq:comp_naive} and \eqref{eq:traffic_naive}.
The computation cost and traffic can be huge since the number of clients are expected to be large in mobile networks and the data size of the model could be huge, especially when using deep neural networks.

Therefore, we require an intuitive and reasonable, but more computation-and communication-efficient metric to evaluate clients' contributions.

\vspace{3mm}
\noindent \textbf{Step-wise contribution} \\
We propose a light-weight but intuitive contribution estimation method based on step-wise contribution calculation.
The metric used in the proposed method is defined as the sum of gains that include the client model in each round, calculated as:
\begin{equation}
    G^{\mathrm{SWC}}_i = \frac{\sum_{r = 1}^{r_\mathrm{end}}P(M^{(r)}_\mathcal{C}) - P(M^{(r)}_{\mathcal{C} \setminus \{ i \}})}{\sum_{i \in \mathcal{C}} \sum_{r = 1}^{r_\mathrm{end}}P(M^{(r)}_\mathcal{C}) - P(M^{(r)}_{\mathcal{C} \setminus \{ i \}})},
\end{equation}
where the denominator is used for normalization. The metric evaluates how much the client's model improves the global model at each round and regards the sum of the step-wise contributions as the contribution of the FL client. This is based on the intuition that a client that improves model performance at each round will also contribute to the improvement of the final overall model performance.

Note that $G^{\mathrm{SWC}}_i$ cannot be the same as $G^{\mathrm{Naive}}_i$ since $M^{(r)}_j$ used to obtain $M^{(r)}_{\mathcal{C} \setminus \{ i \}}$ is calculated from $M^{(r-1)}_\mathcal{C}$ that involves the effect of $M^{(r-1)}_i$ while $(M_{\mathcal{C} \setminus \{ i \}})$ never involves the effect of client $i$. However, the proposed metric can be calculated in $r_\mathrm{end}$ FL training rounds, \textit{i.e.} the FL operator does not need the additional FL training rounds while the Naive metric requires $(c+1)\cdot r_\mathrm{end}$ FL training rounds. The proposed method requires an increase in model aggregations, model validations and gain calculations by a  factor $c \cdot r_\mathrm{end}$ compared to the original FL training process.
The computation and traffic overhead caused by the proposed method is described as 
\begin{eqnarray}
    \mathrm{Comp.:}&\; & r_\mathrm{end} \cdot c \cdot C_\mathrm{server}, \label{eq:comp_step}\\
    \mathrm{Traffic:}&\; & 0.\label{eq:traffic_step}
\end{eqnarray}
The additional computation, that is $(c-1)$-times model aggregation, is caused at the server at each round. The calculation is completed by the server, and no traffic overhead is caused. Comparing with the naive metric, the computation overhead is reduced by
\begin{equation}
    \frac{c \cdot C_\mathrm{server}}{c\cdot C_\mathrm{server} + \sum_{i\in\mathcal{C}} \sum_{j \in \mathcal{C}\setminus \{i\}}C^{(j)}_\mathrm{ud}}.
\end{equation}
Generally, because the computation power of the server is much higher than those of clients, the additional computation of the proposed method is not a critical issue.


Besides incentive/reward selection, the proposed metric can also be used for other mechanisms such as \textit{e.g.,} a client selection problem where the FL operator selects a subset of clients to participate in a FL round for reducing latency for model distribution and uploading with maintaining the improvement of model performance by the FL \cite{nishio2019client}.

\vspace{3mm}
\noindent \textbf{Other heuristic metrics} \\
Based on the intuition that the clients having copious and diverse data improves the global model, two heuristic metrics can be defined:
\begin{eqnarray}
    \mathrm{Heuristic 1:} \quad G_i^\mathrm{H1} = D_i/\sum_{i \in \mathcal{C}}{D_i}, \\
    \mathrm{Heuristic 2:} \quad G_i^\mathrm{H2} = D_i\cdot v_i/\sum_{i \in \mathcal{C}}{D_i\cdot v_i},
\end{eqnarray}
where $D_i$ denotes the number of data sample stored by client $i$. $v_i$ denotes the varieties of the data owned by client $i$, which refers specifically to the number of classes of the data samples stored by client $i$ in classification tasks and the range of the target values in the regression tasks.
These metrics are intuitive and easy to calculate. Furthermore, it can be obtained prior to FL training process. However, the metrics do not work in some cases. Supposing that there are three clients having the same number of data with four classes at each client, the client's data class differs from those of the others. The other clients' data classes overlap somewhat. In this case, the heuristic metrics become the same value for each client. However, their actual contributions may differ.

\section{Experimental evaluation}
\subsection{Setup}
A total of $c=3$ clients participated in the FL and joined all the FL rounds. The clients trained their local models using their own data, uploading them to a server that aggregated the models into a global model. We adopted an image classification task leveraging MNIST dataset \cite{MNIST}, which is a dataset of handwritten digits and is commonly used as a benchmark. The MNIST dataset consists of 60,000 training images and 10,000 testing images with digits of 0--9 stored as 28x28 pixels. 


The small portions of the training dataset were distributed to $c=3$ clients. The total number of data samples stored by Client 3 was fixed to $D_3=350$, and those of Client 1 and 2 were $D_1, D_2 \in [50, 200, 350]$. We considered a non-i.i.d setting, wherein each client possessed data samples from specific classes of the 50,000 training samples. 
Specifically, Client 1 stored $D_1$ samples by randomly sampling from $v_1 = 7$ classes of digits (i.e., 0--6). Client 2 stored $D_2$ samples by randomly sampling from $v_2 = 7 \text{ or } 5$ classes of digits (i.e., 0 or 2 to 6). Client 3 stored $350$ samples by randomly sampling from $v_3 = 3$ classes of digits that did not overlap the classes Client 1 (i.e., 7--9). In these settings, Client 1 stored samples of the most various classes ($v_1 \geq v_2 > v_3$), and Client 2 stored digits that were stored by Client 1. The digits of 7--9 were stored by only Client 3. Therefore, the contributions of Client 2 were expected to trend smaller than those of the other clients. The 10,000 testing images were used for the model validation at each round.

We implemented convolutional neural networks as global and local models with TensorFlow \cite{TF}. Specifically, the model consisted of two $3\times 3$ convolution layers of 16 and 32 channels, each of which was activated by a ReLU and batch normalized. Each convolution layer was followed by $2\times 2$ max pooling. The last max pool layer was followed by two fully-connected layers of 64 units with ReLU activation and another 10 units activated by soft-max. 
The batch size, the number of epochs for local training $E$, and the number of FL rounds were set to $B=50$, $E=30$, and 30, respectively. A SGD having a learning rate of 0.25 was used for the optimizer of each client.

We evaluated the errors of the metrics defined in Sect.~\ref{sec:metric} from the Naive contribution metric.
The error was defined as Euclidean distance between the metrics, which is written as follows:
\begin{equation}
    E = \sqrt{(G^{\mathrm{Naive}}_1 - G_1^{*})^2+(G^{\mathrm{Naive}}_2 - G_2^{*})^2+(G^{\mathrm{Naive}}_3 - G_3^{*})^2}.
\end{equation}

\subsection{Results}
\begin{figure}[t]
    \centering
    \includegraphics[width=1.0\linewidth]{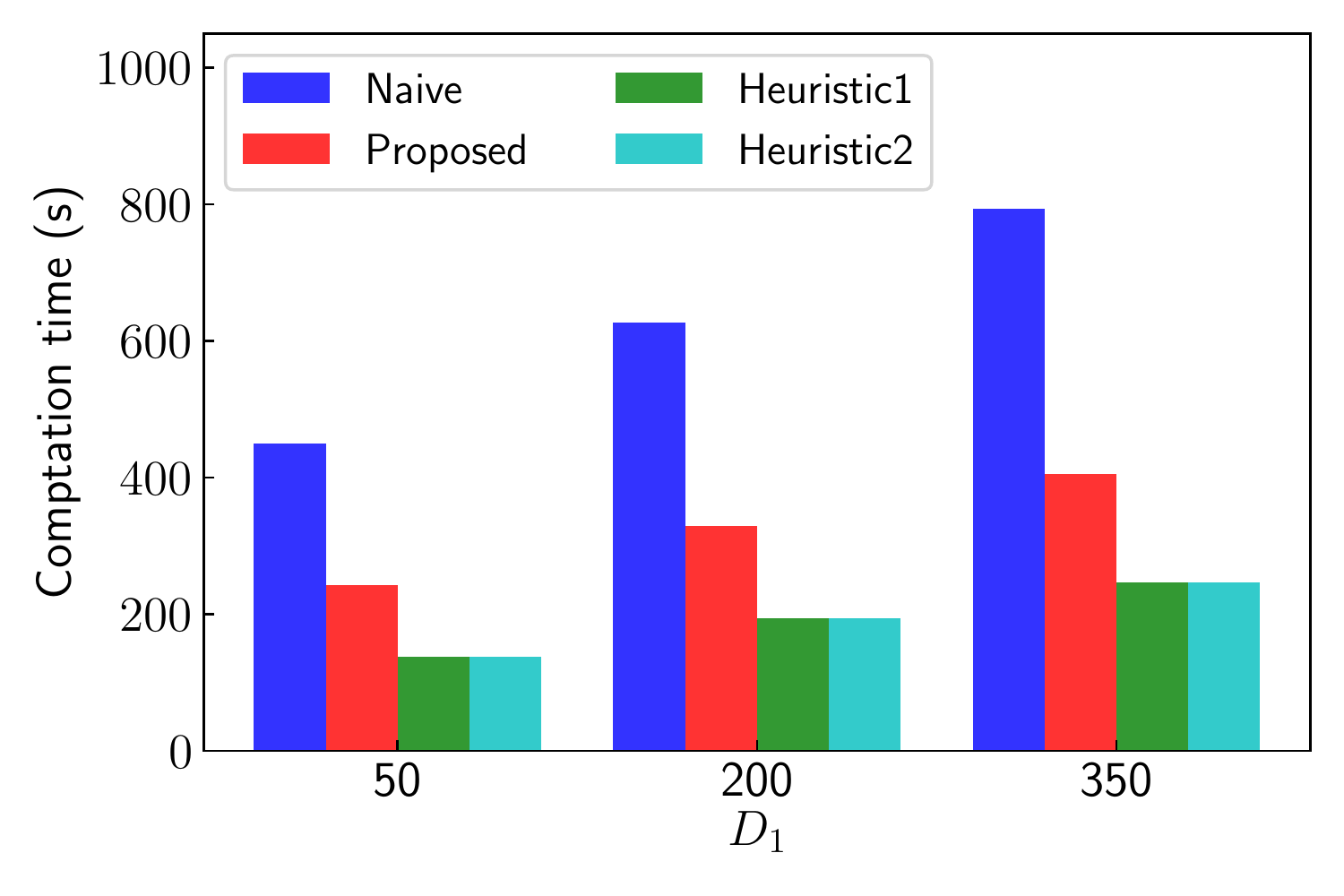}
    \caption{Total computation time for the FL and evaluating contributions.}
    \label{fig:comp_time}
\end{figure}

Figure~\ref{fig:comp_time} depicts a total computation time required for the FL and evaluating contributions, which did not include latency for model transmissions. Thus, this result indicates the computation load of each method. Since the computation load to calculate heuristic metrics was negligible, which was hundreds microseconds in the experiments, the computation time of the heuristic methods were almost the same as the usual FL. On the other hand, the naive and proposed method increased in the computation time. However, the proposed method increases much shorter time than the naive method, which means that the proposed method requires much smaller computation overhead as discussed in Sect.\ref{sec:metric}. 

\begin{figure}[t]
    \centering
    \includegraphics[clip, width=1.0\linewidth]{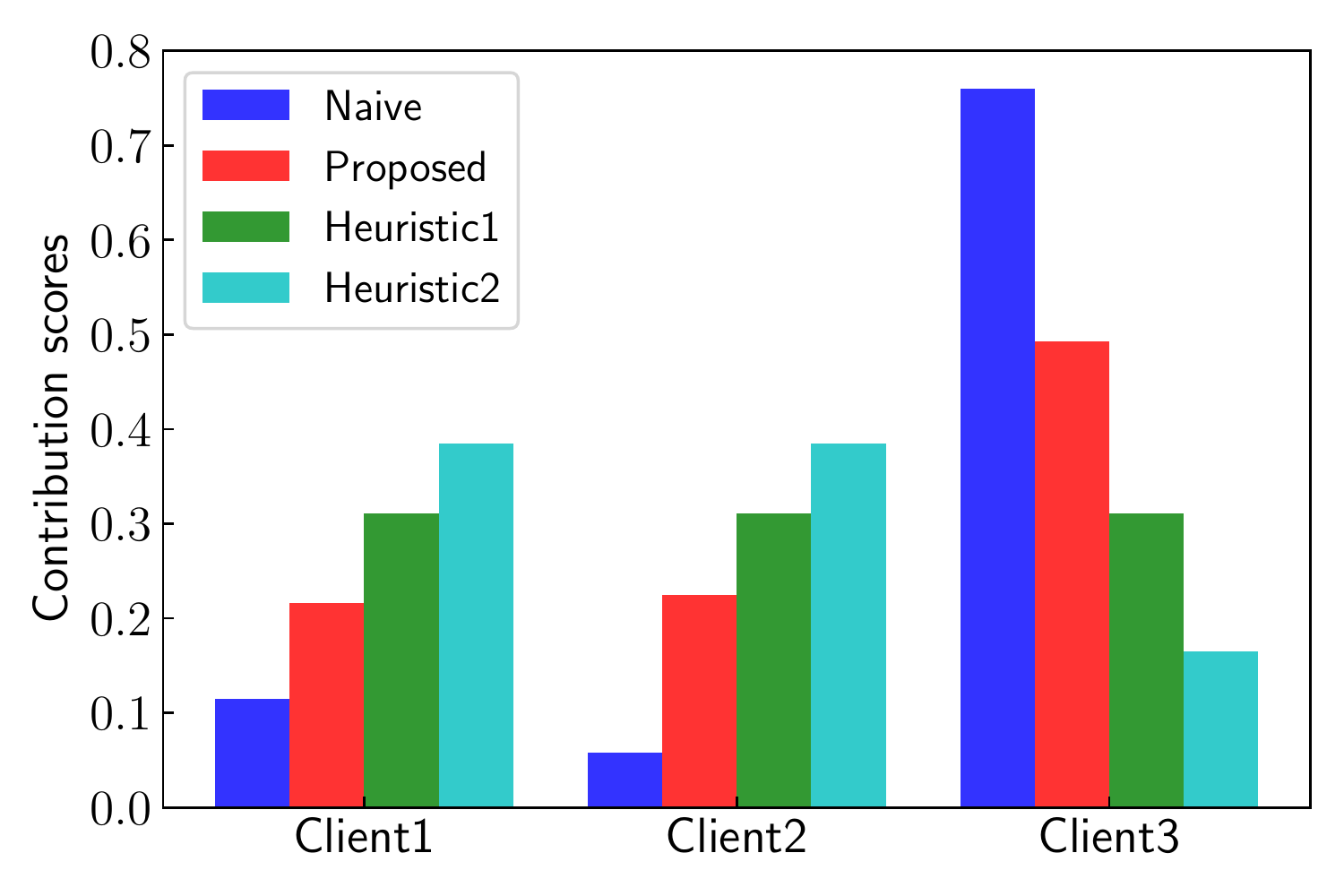}
    \caption{Client contribution scores when Clients 1 and 2 had 350 training samples of 0--6 digits, and Client 3 had 350 samples of 7--9 digits.}
    \label{fig:res_773}
\end{figure}

Figure~\ref{fig:res_773} depicts the contribution scores of each client when Clients 1 and 2 had 350 training samples of 0--6 digits, and Client 3 has 350 samples of 7--9 digits. In the every methods, the contribution scores of Clients 1 and 2 were nearly the same. This is reasonable scoring, because Clients 1 and 2 had the same number and variety of training samples. On one hand, the score of Client 3 was higher than the others according to the naive and proposed methods. This is also reasonable, because the data stored by Client 3 was unique, whereas training data for integers 0--6 were stored at both clients. When Client 3 left the FL, the global model could not achieve an accuracy higher than 0.7. The model could, however, achieve greater than 0.7 if the Clients 1 or 2 left, because both had training data of 0--6 digits. The heuristic methods for Clients 1 and 2 did not consider the data uniqueness and gave the same or lower score to Client 3, which is unreasonable in this scenario.

\begin{figure}[t]
    \centering
    \includegraphics[width=1.0\linewidth]{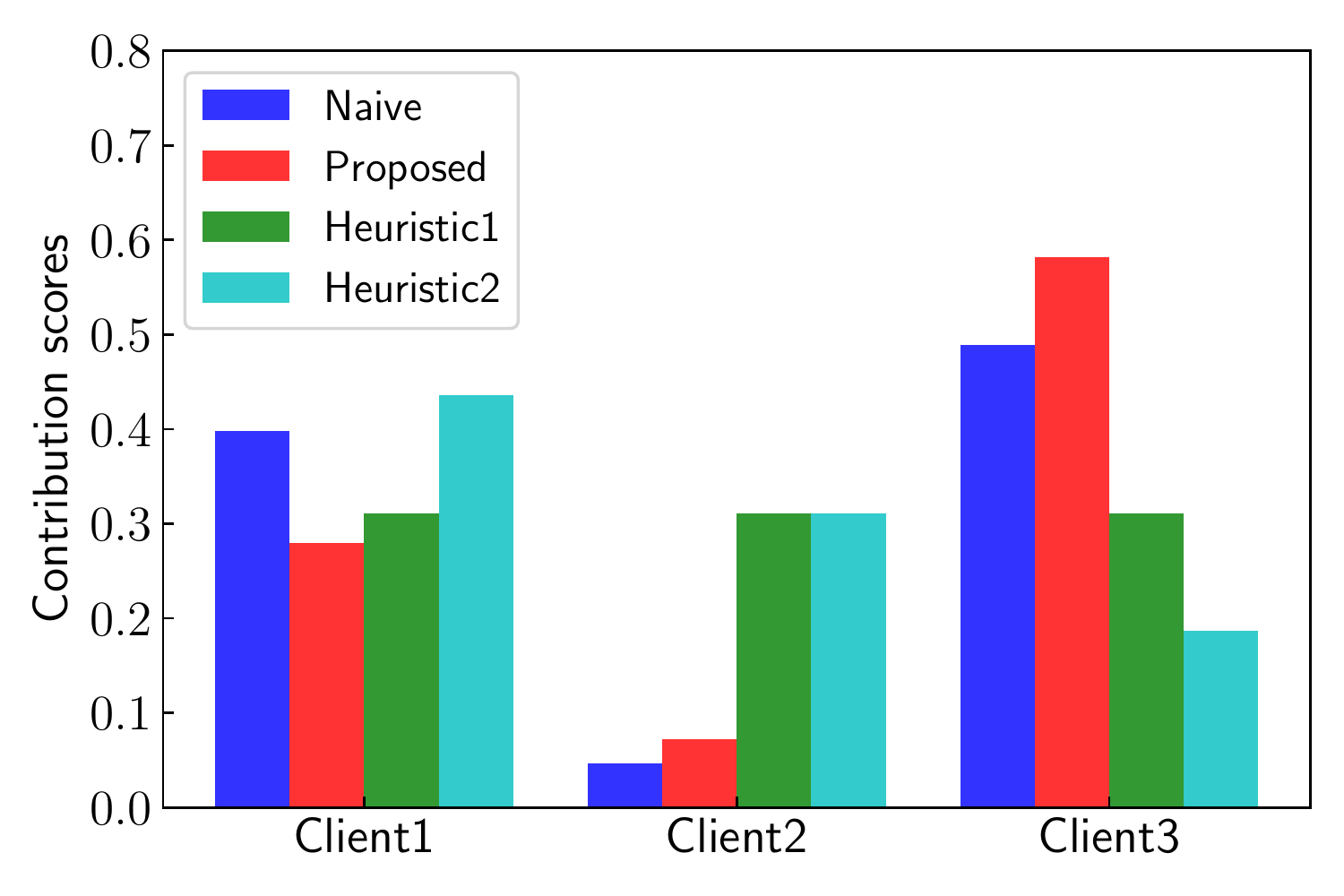}
    \caption{Client contribution scores when Clients 1,2 and 3 had 350 training samples of 0--6 digits, 2--6 digits, and 7--9 digits, respectively.}
    \label{fig:res_753}
\end{figure}

Figure~\ref{fig:res_753} depicts the contribution scores of each client when Clients 1,2, and 3 had 350 training samples of 0--6 digits, 2--6 digits, and 7--9 digits, respectively. In this setting, the contributions of Client 2 must have been smaller than Client 1, because Client 2 had a smaller variety of training samples. The naive and proposed methods gave the lower score to Client 2 as expected, whereas the heuristic methods did not.

\begin{figure}[t]
    \centering
    \includegraphics[width=1.0\linewidth]{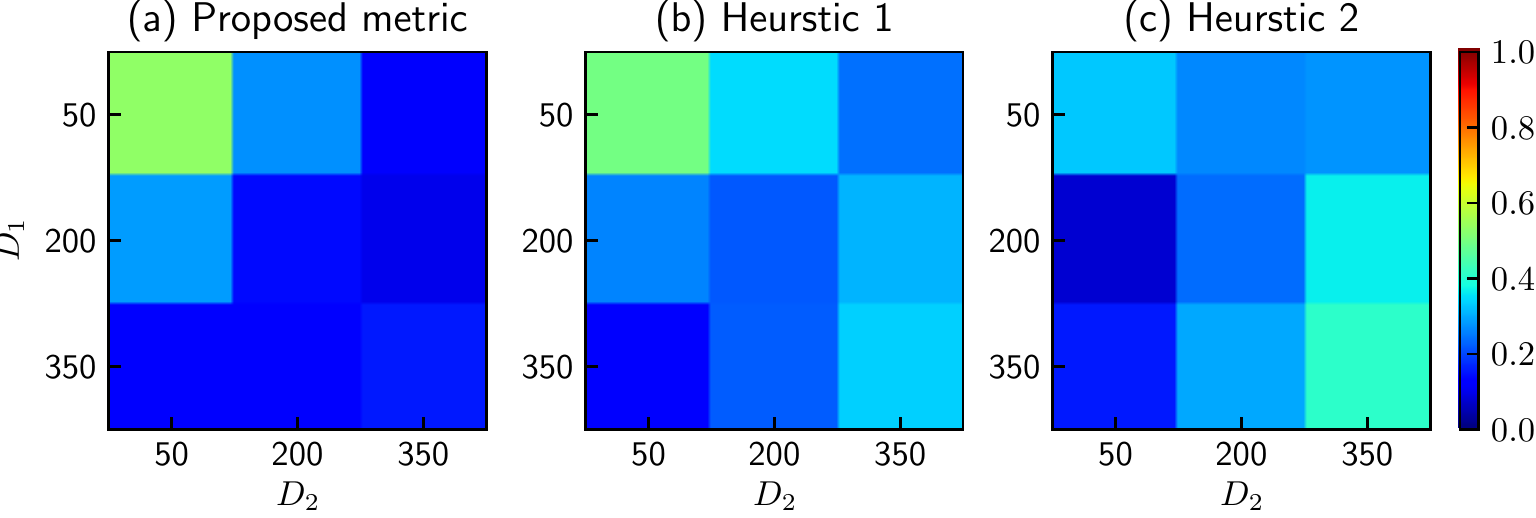}
    \caption{Errors between Naive metric and each method when Client 2 had digits of 5 classes. Average error were Proposed: 1.80, Heuristic1: 2.50, Heuristic2: 2.36.}
    \label{fig:res_1}
\end{figure}

Figure~\ref{fig:res_1} depicts that errors of scores from the naive method when changing the data amount of Clients 1 and 2, $D_1, D_2$. Here, Clients 1 and 2 had training samples of 0--6 and 2--6 digits, respectively. The proposed metric achieved a smaller gap from the naive method when the training samples of Clients 1 and 2 became larger, whereas the gaps in the other methods became larger. 

\begin{figure}[t]
    \centering
    \includegraphics[width=1.0\linewidth]{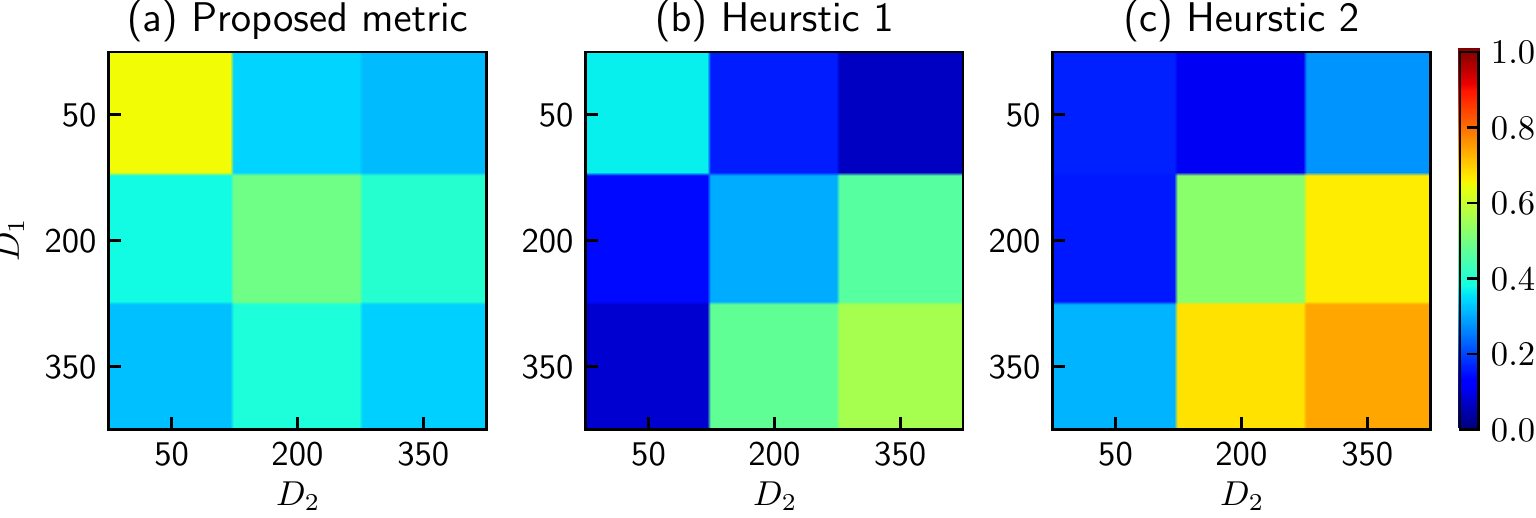}
    \caption{Errors between Naive metric and each method when Client 2 had digits of 7 classes. Average error were Proposed: 3.57, Heuristic1: 2.55, Heuristic2: 3.55}
    \label{fig:res_2}
\end{figure}

Figure~\ref{fig:res_2} depicts that the errors that occurred when Clients 1 and 2 had the same classes of training samples. This result shows cases in which the proposed method achieved larger errors than other methods.
When $D_1$ and $D_2$ were larger than 200, the proposed method achieved lower errors than other methods as with the results shown in Figs.~\ref{fig:res_773}--\ref{fig:res_1}. On the other hand, when $D_1$ or $D_2$ was 50, the error of the proposed method became larger than other methods. When $D_1$ or $D_2$ was 50, the proposed method estimated that the contribution of the client with 50 data samples was less than other clients but not zero, while the naive method estimated that its contribution was almost zero, which made the larger gap. This is because that the model improvement at each round by a client with small but diverse data samples is little but unstable, which could cause large but temporal model improvement and increase in the step-wise contribution metric employed in the proposed method. This drawback of the proposed method can be mitigated by restricting clients with so small amount of data, which are not expected to improve the global model, from participating the FL.

\section{Conclusion}
This paper proposed an intuitive and computation-and communication-efficient method to estimate the individual contribution levels of participants in FL so as to determine appropriate incentive mechanisms for participation in FL. 
The proposed method enabled the evaluation during a single FL training process, there by reducing the need for traffic and computation overhead. The performance evaluations using the MNIST dataset showed that the proposed method estimated the contributions of the individual clients with much smaller computation and communication overhead than those of the naive method.

\bibliographystyle{IEEEtran}
\bibliography{reference.bib}

\end{document}